\documentclass[letterpaper, 10 pt, conference,twoside]{ieeeconf}  

\IEEEoverridecommandlockouts                              

\usepackage{amsmath,amsfonts}
\usepackage{algorithmic}
\usepackage{array}
\usepackage[caption=false,font=normalsize,labelfont=sf,textfont=sf]{subfig}
\usepackage{textcomp}
\usepackage{stfloats}
\usepackage{verbatim}

\usepackage{xcolor}
\usepackage{color}
\usepackage{soul}

\usepackage{graphicx}
\usepackage{svg}
\usepackage{dirtree}

\usepackage{hyperref}
\usepackage{url}
\usepackage{float}
\usepackage{dirtytalk}
\usepackage[capitalise]{cleveref}
\usepackage{fancyhdr}
\usepackage{lastpage}
\usepackage{booktabs,caption}
\usepackage{listings}
\usepackage{csquotes}

\usepackage{amssymb}
\usepackage{pifont} 

\usepackage{tablefootnote} 
\usepackage{multirow}
\usepackage{bm,array}
\usepackage{longtable}

\usepackage{romannum}
\AtBeginDocument{\pagenumbering{arabic}}

\definecolor{codegreen}{rgb}{0,0.6,0}
\definecolor{codegray}{rgb}{0.5,0.5,0.5}
\definecolor{codepurple}{rgb}{0.58,0,0.82}
\definecolor{backcolour}{rgb}{0.95,0.95,0.92}

\lstdefinestyle{mystyle}{
    backgroundcolor=\color{backcolour},   
    commentstyle=\color{codegreen},
    keywordstyle=\color{magenta},
    numberstyle=\tiny\color{codegray},
    stringstyle=\color{codepurple},
    basicstyle=\ttfamily\footnotesize,
    breakatwhitespace=false,         
    breaklines=true,                 
    captionpos=b,                    
    keepspaces=true,
    frame=single,
    numbers=none,                    
    numbersep=5pt,                  
    showspaces=false,                
    showstringspaces=false,
    showtabs=false,                  
    tabsize=2
}

\lstdefinelanguage{CheckUrdf}
{
    morekeywords={check_urdf},
}

\definecolor{orgred}{rgb}{0.8078,0.4471,0.2314}
\definecolor{darkgreen}{rgb}{0.4157,0.6,0.333}
\definecolor{darkblue}{rgb}{0.0,0.0,0.6}
\definecolor{cyan}{rgb}{0.0,0.6,0.6}
\definecolor{light-gray}{gray}{0.80}

\lstdefinestyle{xmlStyle}{
  basicstyle=\ttfamily\scriptsize,
  columns=fullflexible,
  showstringspaces=false,
  commentstyle=\color{darkgreen},
  numbers=left,                    
  numbersep=10pt, 
  numberstyle=\tiny,
  captionpos=b,
}
\lstdefinelanguage{XML}
{
  morestring=[b]",
  morestring=[s]{>}{<},
  morecomment=[s]{<?}{?>},
  stringstyle=\color{black},
  identifierstyle=\color{darkblue},
  keywordstyle=\color{cyan},
  morekeywords={xmlns,type,name,xyz,link,size,radius,length},
}

\definecolor{orange}{rgb}{1,0,0}
\definecolor{grey}{rgb}{0.5,0.5,0.5}

\newif\ifcomments %

\commentstrue 

\usepackage{fancyhdr}

\title{\textbf{Understanding URDF: A Survey Based on User Experience}}

\author{Daniella~Tola and Peter~Corke~\IEEEmembership{Fellow,~IEEE}%
\thanks{D. Tola is with the Department of Electrical and Computer Engineering, Aarhus University, Aarhus, Denmark (e-mail: dt@ece.au.dk).}%
\thanks{P. Corke is with the Queensland University of Technology (QUT) Centre for Robotics, QUT, Brisbane, Australia (e-mail: peter.corke@qut.edu.au).}%
\thanks{This work was supported by the Innovation Foundation Denmark through the MADE FAST project. 
We would like to thank our colleagues for the detailed feedback on the survey, and particularly Cláudio Gomes for helping with both revision of the questions and distribution of the survey. 
The study was approved by the Institutional Review Board at Aarhus University with the approval number TECH-2022-013.}
}

\begin{document}

\maketitle
\thispagestyle{empty}
\pagestyle{empty}

\begin{abstract}
With the increasing complexity of robot systems, it is necessary to simulate them before deployment.
To do this, a model of the robot's kinematics or dynamics is required.
One of the most commonly used formats for modeling robots is the Unified Robot Description Format (URDF).
The goal of this article is to understand how URDF is currently used, what challenges people face when working with it, and how the community sees the future of URDF.
The outcome can potentially be used to guide future research.

This article presents the results from a survey based on 510 anonymous responses from robotic developers of different backgrounds and levels of experience.
We find that 96.8\% of the participants have simulated robots before, and of them 95.5\% had used URDF.
We identify a number of challenges and limitations that complicate the use of URDF, such as the inability to model parallel linkages and closed-chain systems, no real standard, lack of documentation, and  a limited number of dynamic parameters to model the robot.
Future perspectives for URDF are also determined, where 53.5\% believe URDF will be more commonly used in the future, 12.2\% believe other standards or tools will make URDF obsolete, and 34.4\% are not sure what the future of URDF will be.
Most participants agree that there is a need for better tooling to ensure URDF's future use.
\end{abstract}

\section{Introduction}

Simulating robots has become an integrated part of robot system development~\cite{Sannemann&2020,Afzal&2021}, as it reduces the cost by allowing experimentation with parameters and environments in advance of committing to the physical hardware.
A commonly used method to represent a robot's geometry and physical appearance is the Unified Robot Description Format (URDF)~\cite{wikiROS}.
It was introduced with the Robot Operating System (ROS) in 2009 as a format to describe the kinematics, geometries, and dynamics of robots in a universal manner~\cite{Quigley&2015}.

A URDF file is an XML-based file with an extension of \emph{.urdf}, that can be imported and exported by different tools for visualization or simulation purposes.
It describes robot links (rigid bodies) and the joints that connect the links using a tree structure.

\begin{figure}
    \centering
    \includegraphics[width=\columnwidth]{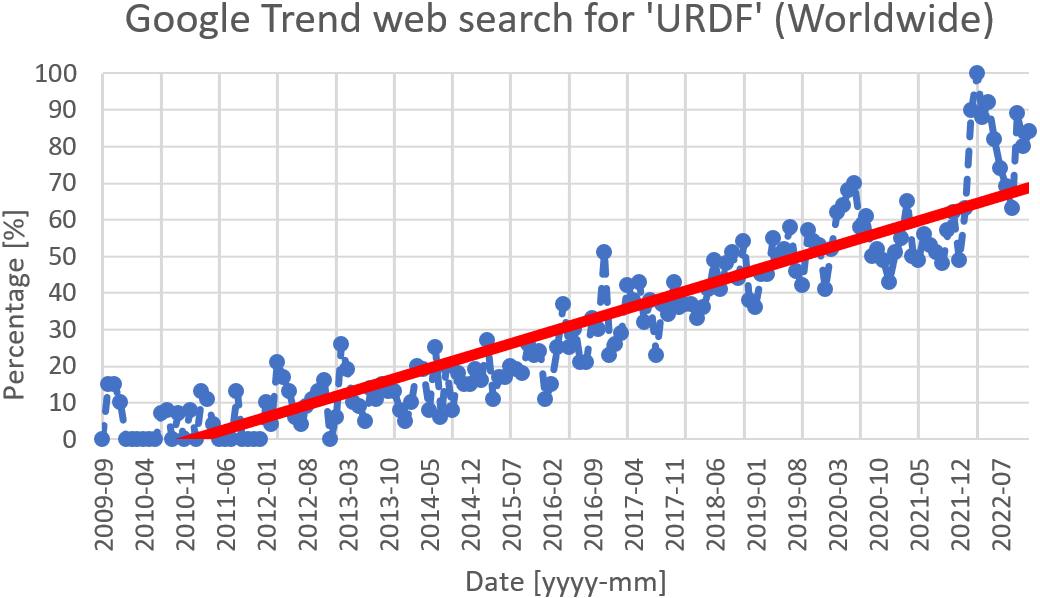}
    \caption{Google trend (\url{trends.google.com/trends}) for the search term `URDF' since 2009. Note that the data is an approximation of the number of Google searches. The data points where the search is zero percent are due to insufficient data.}
    \label{fig:trend_urdf_oct2009}
\end{figure}

\begin{figure}[!b]
    \centering
    \includegraphics[width=0.6\columnwidth]{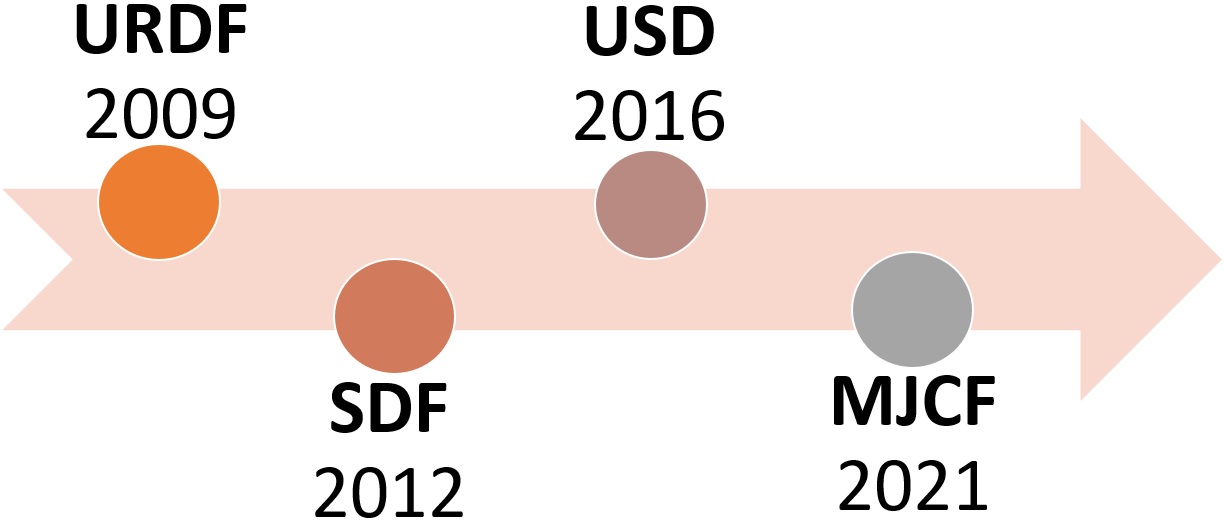}
    \caption{Timeline presenting the year of the first commit to the default branch of the GitHub repository of each description format.}
    \label{fig:timeline}
\end{figure}

The use of URDFs has increased over the years, as evidenced by the trend in Google queries for the term `URDF', see \cref{fig:trend_urdf_oct2009}.
Although URDF is widely used, it has inherent issues which have been made clear by multiple roboticists through online forums such as ROS Discourse~\cite{discourseRobotDescriptionFormats} (posts from 2016-2022), GitHub repositories~\cite{githubURDF2Sachin} (posted in 2015), and Google groups~\cite{googleGroupsURDF} (posted in 2015).
Additionally, the first ROSCast (in 2016) at MetroRobots.com discussed some of the underlying issues of URDF and potential future directions~\cite{metroRobotsROSCast}.
At the time of writing (February 2023) many of the challenges with URDF, that were described since 2015, still persist.
Over time, see \cref{fig:timeline}, other open-source object description formats have been developed such as the Simulation Description Format (SDF)~\cite{SDFformat}, Universal Scene Description (USD)~\cite{USDformat}, and MuJoCo Modeling XML File (MJCF)~\cite{MJCFormat}.

With various formats emerging, it is necessary to understand the current state of URDF and its potential future directions.
We conducted an anonymous survey on the experience of roboticists with URDF, where we asked about the development of URDF, its limitations, and desired improvements.
The results of this survey can be used to direct future research on URDF and robot description formats.

Our contributions in this article include:
\begin{itemize}
    \item a study with over 500 robotics developers,
    \item an analysis of challenges and possible improvements,
    \item presentation of quantitative and qualitative data on the current use of URDF and perspectives for its future use,
    \item experienced users' knowledge of tools to create URDFs,
    \item and the survey materials and results allowing other roboticists to build on top of this work.
\end{itemize}

\Cref{sec:background} provides a brief introduction to URDF, followed by the methodology of the survey in \cref{sec:methodology}, and the results in \cref{sec:results}.
An overall summary and discussion of the survey results is presented in \cref{sec:discussion}.
We conclude in \cref{sec:conclusion} with our main findings and directions for future work.
\section{Background} \label{sec:background}

\begin{figure}
    \centering
    \includegraphics[width=\columnwidth]{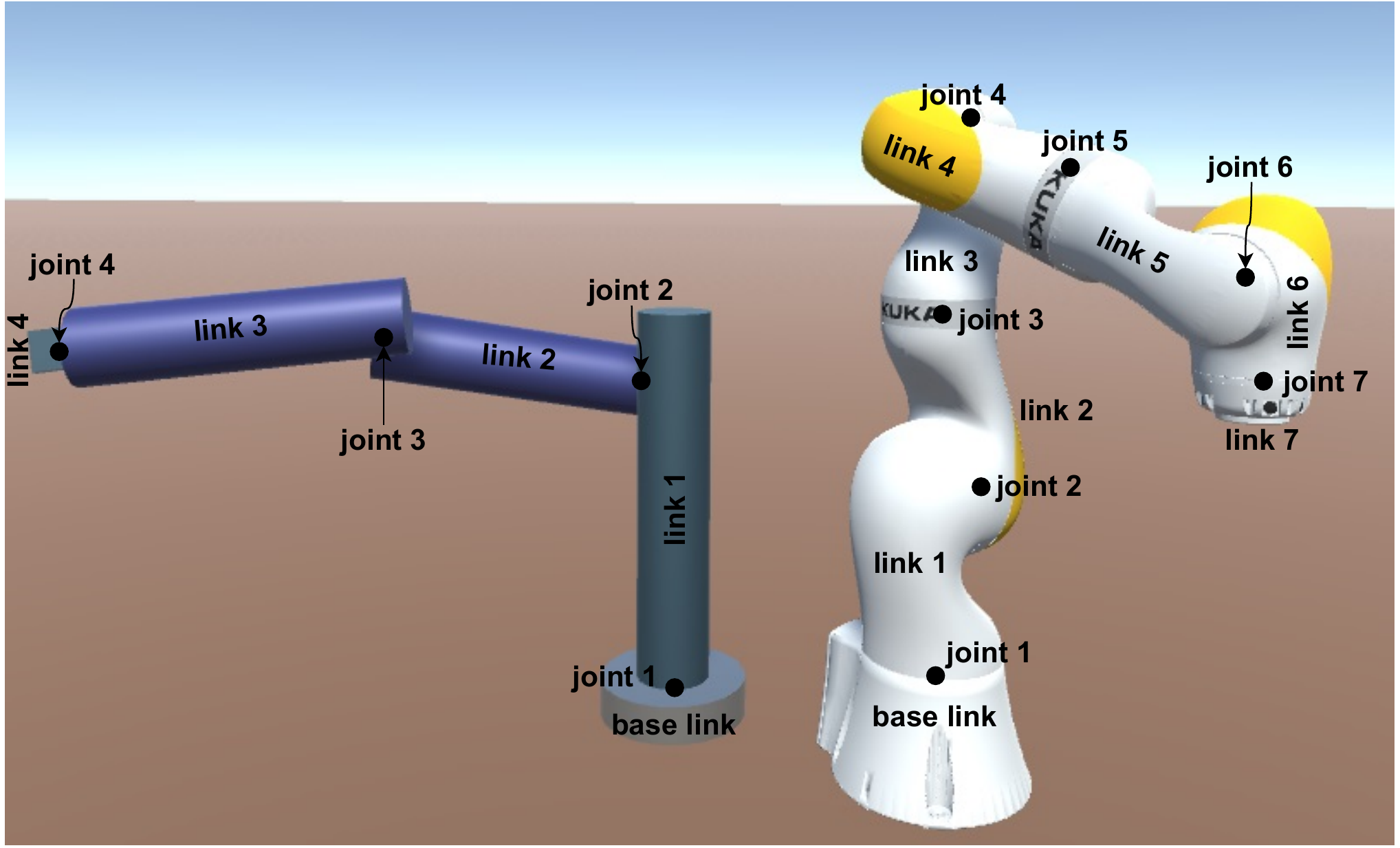}
    \caption{Example of a URDF with basic shapes (left) and a URDF with polygon meshes (right).}
    \label{fig:cylindrical_vs_kuka}
\end{figure}

\noindent A URDF file is a human-readable XML file describing the kinematic structure, dynamics, and visual representation of a robot.
The visual representation can be defined using basic 3D shapes such as boxes and cylinders or using 3D polygon meshes that typically consist of connected triangles defining the shape of the object.
\Cref{fig:cylindrical_vs_kuka} shows two examples of robotic arms modeled with URDF, the left one using basic 3D shapes, and the right one using 3D polygon meshes.

\subsection{URDF File} \label{subsec:urdf_urdf_file}

URDF is a stand-alone file format that includes relevant modeling details of a robot.
It was initially introduced with ROS but has since been adopted by tools both within and outside of the ROS ecosystem.
Listing \ref{lst:urdf2dof} illustrates how a link and a joint of a robot can be modeled using URDF.

\definecolor{baselinkcolor}{rgb}{0.96,0.88,0.76}
\definecolor{joint1color}{rgb}{0.96,0.93,0.89}
\definecolor{others}{rgb}{0.76,0.74,0.82}
\newcommand{\coloropacity}{!65}%

\newcommand{\Hilight}[1]{\makebox[0pt][l]{\color{#1}\rule[-4pt]{0.95\columnwidth}{10pt}}} 
 \begin{minipage}{0.9\columnwidth} 
\begin{lstlisting}[caption={Parts of a URDF file for the robotic arm shown on the left in \cref{fig:cylindrical_vs_kuka}.},label={lst:urdf2dof},style=xmlStyle]
|\Hilight{baselinkcolor\coloropacity}| <link name="base link">
|\Hilight{baselinkcolor\coloropacity}\hspace{0.2cm}|  <visual>
|\Hilight{baselinkcolor\coloropacity}\hspace{0.4cm}|   <origin xyz="0 0 0.25"/>
|\Hilight{baselinkcolor\coloropacity}\hspace{0.4cm}|   <geometry>
|\Hilight{baselinkcolor\coloropacity}\hspace{0.65cm}|    <cylinder length="0.05" radius="0.1"/>
|\Hilight{baselinkcolor\coloropacity}\hspace{0.4cm}|   </geometry>
|\Hilight{baselinkcolor\coloropacity}\hspace{0.2cm}|  </visual> 
|\Hilight{baselinkcolor\coloropacity}| </link>
...
|\Hilight{joint1color\coloropacity}| <joint name="joint 1" type="continuous">
|\Hilight{joint1color\coloropacity}\hspace{0.2cm}|  <parent link="base link" />
|\Hilight{joint1color\coloropacity}\hspace{0.2cm}|  <child link="link 1" />
|\Hilight{joint1color\coloropacity}\hspace{0.2cm}|  <axis xyz="0 0 1" />
|\Hilight{joint1color\coloropacity}\hspace{0.2cm}|  <origin xyz="0 0 0.05"/>
|\Hilight{joint1color\coloropacity}| </joint>
\end{lstlisting}
\end{minipage}

\subsubsection{Links}
are rigid bodies of the robot that can be connected using joints.
The {\small{\texttt{base link}}}, on lines 1-8 in Listing \ref{lst:urdf2dof}, represents the fixed base of the robot which is visualized as a cylinder, the upright gray cylinder shown in \cref{fig:cylindrical_vs_kuka} (left robot).
Further information on link properties can be found in the ROS Wiki page~\cite{wikiROSLink}.
Note, that only links of rigid bodies can be represented using URDF, deformable bodies cannot.

\subsubsection{Joints}
connect two links, a parent and child link.
The parent link is closer to the robot base, and the child link is closer to the tool tip.
The joint, on lines 10-15 in Listing \ref{lst:urdf2dof}, connects the {\small{\texttt{base link}}} with another link, {\small{\texttt{link 1}}}.
These two links can be seen on the left robot in \cref{fig:cylindrical_vs_kuka}, in which the bottom light gray cylindrical link is the parent link, and the long vertical dark gray cylindrical link is the child link.
This joint is continuous, meaning it is revolute with no motion limits.
Additional joint properties can be found in the ROS Wiki page~\cite{wikiROSJoint}.


\subsection{Xacro}
Xacro is a macro language for XML that is used to construct URDF files~\cite{wikiROSXacro} and the xacro preprocessor is part of the ROS ecosystem.
It provides tags that can be used to configure URDF files based on the application, to reduce redundancy and improve the maintainability of the models.
These tags are defined in a URDF-based xacro file with the extension \emph{(.xacro)}.
The xacro preprocessor is an executable used to generate URDF files by interpreting the xacro tags and using input values.
The preprocessor allows combining multiple xacro files, which is especially beneficial when dealing with complex robotic structures~\cite{Albergo&2022}.
\section{Methodology} \label{sec:methodology}


\subsection{Motivation}
The motivation for creating the survey is to understand the current position of URDF in the robotics community.
We use the survey to answer the following questions:

\begin{itemize}
    
    \item[] \textbf{Q1:} Who is using URDF?

    \item[] \textbf{Q2:} What is URDF being used for?

    \item[] \textbf{Q3:} What are the limitations of URDF?

    \item[] \textbf{Q4:} What is the future of URDF?

\end{itemize}

We ask these questions to understand how common the format is, its perceived complexity by the robotics community, and how users envision its future.
These questions are important to guide the direction of future research in the area of robot description formats.


\subsection{Survey Design}

An online survey among robotics developers was conducted in December 2022 to answer our research questions.
The survey contained in total 22 questions, of them 3 are demographic, 16 are multiple choice where 8 of them also had a text option, 2 are rating questions, and 1 is open-ended.
The demographic questions were used to understand the experience and background of the participants.
This structure of starting with demographic questions was adapted from another survey on simulation for testing robots~\cite{Afzal&2021}.
We used the online tool \textit{SurveyXact}~\cite{SurveyXact} to conduct the survey and collect data.
To avoid asking irrelevant questions, the survey was structured with different paths depending on the participant responses, resulting in varying numbers of responses for different questions.
A list of the questions, their ID, and the number of responses is presented in the appendices in \cref{tab:questions_ids_responses}.
We will refer to the questions by their ID.
Ethical approval of the survey was granted by the Institutional Review Board at Aarhus University with the approval number TECH-2022-013.


\subsection{Recruitment}

To collect responses we distributed the survey via email to the robotics community, through LinkedIn posts (by both authors, nearly 40.000 impressions), a Twitter post, an email to robotics-worldwide, a ROS Discourse post, and posting in robotics Reddit communities.
In total, 689 participants took the survey, where 515 of them fully completed the survey and the remaining partially completed it.
Out of the 515 responses, 5 did not give their consent therefore ending the survey, leaving 510 completed responses which this article is based on.
In our recruitment emails and posts, we described that we are conducting a survey on robot simulation and URDF.
As URDF was directly stated in the recruitment description, we can assume that most participants have experience with or knowledge of the format.

The experience of the participants and the organizations where they have simulated robots or worked with URDF are presented in \cref{tab:demographics_simulation}, where 13/510 participants had no experience with this.
In total 16/510 participants had never simulated a robot before (based on answers to D1 and S1).
The majority of the participants have heard of URDF (477/497) and used it (472/477).
The experience of the participants on URDF is presented in \cref{tab:demographics_urdf}.
Participants reported different levels of experience with simulation and URDFs, within different sizes and types of organizations.
We can claim that the sample size (510) and the participants' background is diverse ensuring that the study is not limited to one type of population, but there is a potential bias towards people in academia (82\%) compared to industry (52\%).


\subsection{Analysis}

The survey consisted of both quantitative and qualitative questions, allowing the participants to express their own opinion on the challenges they found, desired improvements, and future perspectives on URDF.
We used descriptive coding to analyze the open-ended questions, which entails summarizing responses into words or phrases that describe a main topic in the data.
Specifically, we used \textit{in vivo coding} as the phrases defining the topics are taken directly from terms used by the participants themselves~\cite{Saldana2009}.
The classification of responses into the different topics was performed based on our subjective opinion.


\subsection{Threats to Validity}

To mitigate the risk of bias in the questions, we followed survey best practices~\cite{Kitchenham&95} such as allowing open-ended answers for participants to explain their choice, and iteratively pre-testing with volunteers.

Typical biases that can occur in multiple-choice questions are position bias and forced-choice bias.
Position bias can occur depending on the order of the choices, and to reduce this we have ordered the choices alphabetically.
Forced-choice bias occurs when participants do not agree with any of the choices, but are forced to choose one of the options to continue the survey.
To reduce this, we have in many questions allowed participants to select ``Other'' and describe their answer using text, allowing them to provide information based on their subjective experience/opinion.
A bias may also occur when allowing participants to answer both via multiple-choice and text, as many participants want to limit the time spent on a question, and therefore may choose to skip providing additional information using the text option.

Adding further comments or notes was possible at the end of the survey.
Some of these comments can be found in the appendix in \cref{app_results}, and pointed out potential biases in the formulation of the questions, where:
\begin{itemize}
    \item Three reported that S10 is limiting and induces a bias (participants A, B, and C). Question 10 asks the participants how they have used URDF, with the possibility to choose only one of the choices: \textit{`Always in combination with ROS'}, \textit{`Always in combination with ROS, but would prefer not to require ROS'}, or \textit{`Without ROS'}. We agree with the participants' statements and therefore chose to limit the conclusions drawn from these results.
    \item One reported S16 is leading by using the word \textit{painful} and makes it clear that we are looking for a specific result (participant D). The question is open-ended and goes as follows \textit{`What would you say is the most painful part of creating/modifying a URDF model?'}. One of our motivations for creating this survey is a result of our experience with the workflow of creating URDFs, which we believe is painful, and is unfortunately made clear in the question potentially inducing a bias.
    \item One reported S18 as biased, as it seemed we were asking which of the following improvements should we implement first (participant A).
    This does not reflect the truth behind the question and its choices.
    The improvements we listed in the multiple-choice question were found on community posts and were internally discussed and agreed upon as what evidently seemed to be desired by people.
    Naturally, such a question will be biased as the majority will choose answers from the list instead of writing their own opinion.
\end{itemize}

External validity describes how well the results can be generalized~\cite{Ghazi&2019}.
The threat to external validity in this survey is reduced with a large number of participants ($>$500), their various organizational backgrounds, and differing years of experience, see \cref{tab:demographics_simulation} and \cref{tab:demographics_urdf}.

Construct validity is related to generalization of results with focus on correctly labeling results by making inferences based on observations.
As we have categorized the responses to open-ended questions ourselves, there is a potential threat to the construct validity in the results.
Conclusion validity concerns defining the relationship between variables in the data and whether or not the conclusion is reasonable.
As we have compared variables between responses from different questions, there is a potential that the conclusions we have made are not entirely reasonable.
To mitigate these threats, we have shared the preliminary findings of the survey with people from the community through the same media used for recruitment.

To promote further research we share our recruitment materials, questionnaire, and additional results at the URL\footnote{\url{github.com/Daniella1/urdf_survey_material}}.

\section{Results and Analysis} \label{sec:results}


\subsection{General Results}

The participants' experience with simulation and/or URDF, and the organization types where they have used this, are shown in \cref{tab:demographics_simulation}.
The participants had last simulated a robot (S1, 497 responses) within the last month (70\%), within the last year (24\%), in the last 5 years (4\%), and 1\% more than 5 years ago.
The remaining 1\% had never simulated a robot.

\begin{table}
\caption{Demographics of the participants' experience with robot simulation and/or URDF. Examples of \textit{Unaffiliated groups} are hobby or school clubs. Note that the tables are positioned independently of each other.}
\label{tab:demographics_simulation}
\begin{tabular}{|p{1.3cm}r|p{1.5cm}r|p{1.5cm}r|}
\hline
\textbf{Years of experience} & \textbf{\%} & \textbf{Organization type} & \textbf{\%} & \textbf{\#Organization employees} & \textbf{\%} \\ \hline
$<$1 & 10.0 & Academia & 82.3 & 1--10 & 26.0 \\ 
1--5 & 61.2 & Industry & 52.5 & 11--50 & 23.7 \\ 
6--10 & 16.7 & Government & 4.0 & 51--100 & 10.3 \\ 
$>$10 & 9.6 & Unaffiliated groups & 14.7 & 100+ & 40.0 \\ 
None & 2.5 & Individual & 25.8 & ~ & ~ \\ 
~ & ~ & Other & 0.4 & ~ & ~ \\ \hline
\multicolumn{2}{|c|}{510 responses} & \multicolumn{2}{c|}{497 responses} & \multicolumn{2}{c|}{493 responses} \\ \hline
\end{tabular}
\end{table}

\Cref{tab:demographics_urdf} shows the participants' experience with URDF.
497/510 participants had simulated a robot or worked with URDF before, where 477 had heard of URDF and 472 had used it.
One of the participants that had not previously used URDF stated it was too difficult, while another stated it was due to the limitations of representing deformable bodies.

\begin{table}
\caption{Participants' experience with URDF. The last column represents how experienced the participants rate themselves with URDF. Note that the tables are positioned independently of each other.}
\label{tab:demographics_urdf}
\begin{tabular}{|p{1.3cm}r|p{1.7cm}r|p{1.4cm}r|}
\hline
\textbf{Years of experience} & \textbf{\%} & \textbf{Last used} & \textbf{\%} & \textbf{Experience} & \textbf{\%} \\ \hline
$<$1 & 16.1 & Last month & 66.5 & Beginner & 21.8 \\
1--5 & 61.8 & Last year & 27.3 & Intermediate & 57.4 \\
6--10 & 15.7 & Last 5 years & 5.3 & Expert & 20.8 \\
$>$ 10 & 5.2 &  $>$5 years ago & 0.8 & ~ & ~ \\
None & 1.0 & ~ & ~ & ~ & ~ \\ \hline
\multicolumn{2}{|c|}{477 responses} & \multicolumn{2}{c|}{472 responses} & \multicolumn{2}{c|}{472 responses} \\ \hline
\end{tabular}
\end{table}

\begin{figure}
    \centering
    \includegraphics[width=\columnwidth]{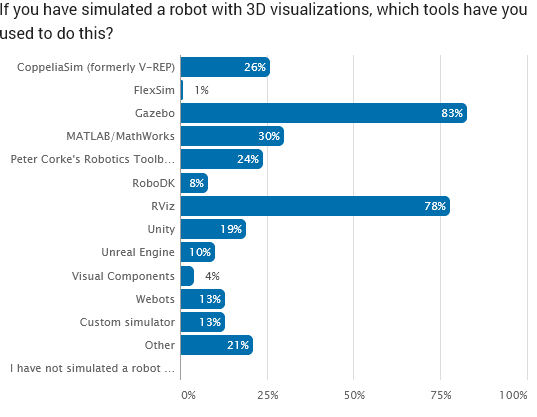}
    \caption{The tools that the participants have used to simulate robots with 3D visualizations (S2, 494 responses).}
    \label{fig:s2_tools_robot_3d_simulation}
\end{figure}

The software tools used by participants to simulate robots (S2, 494 responses) are shown in \cref{fig:s2_tools_robot_3d_simulation}, where Gazebo was the most used tool (83\%), followed by RViz (78\%).
RViz is part of the ROS ecosystem which is provided by the Open Source Robotics Foundation (OSRF)~\cite{OpenRobotics}.
The foundation also provides Gazebo, which is easily integrated to work with the ROS ecosystem.
The least-used tool is FlexSim which less than 1\% of the participants have used.
Of the participants, 21\% stated they use other tools for simulating robots, and the most popular answers among them were PyBullet (17\%), NVIDIA Omniverse consisting of Isaac Sim and Gym (14\%), MuJoCo (11\%), manufacturer-specific tools such as ABB studio or URSim (9\%), and Drake (7\%). 
Manufacturer-specific tools do not typically support URDF, as they provide models of their own robots, making it difficult to simulate a complete application with components from other manufacturers or custom robots.

\subsection{Use of URDF}

\textbf{Application domains (S7, 472 responses):} 
URDF has been used in a variety of application domains, including Manufacturing (46\%), Transportation (24\%), Agriculture (14\%), Medical (12\%), Cleaning (8\%), Defense (6\%), and Marine (6\%).
Additionally, 34\% of respondents stated they used URDF in other domains, with 28\% of the 34\% using them in Academia, 7\% in Service robots (e.g.\ for household, retail, healthcare), 6\% in Logistics or Warehouses, 6\% in Construction robots, and 5\% use URDF in Mining.

\textbf{Manufacturers (S8, 472 responses):} 
Of the robots most commonly used with URDF, the manufacturers are Universal Robots (54\%), Kuka (37\%), Franka Emika (26\%), Robotiq (24\%), ABB (21\%), Clearpath Robotics (21\%), Kinova Robotics (15\%), Willow Garage (14\%), Fanuc (10\%), Boston Dynamics (9\%), and Yaskawa Motoman Robotics (6\%).
In addition, 40\% of the participants stated they use URDFs of other manufacturers, with 31\% stating they have used URDF for self-made/custom robots, and 8\% stating the robot URDFs they use are not of any manufacturer.
The 8\% that did not belong to a manufacturer can also be classified as custom robots, resulting in a total of 39\%.

\begin{figure}
    \centering
    \includegraphics[width=\columnwidth]{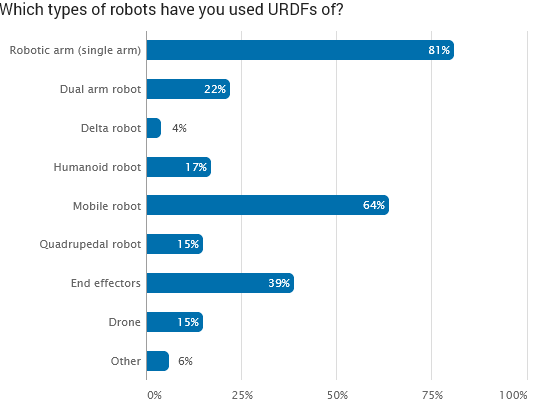}
    \caption{Robot types modeled in URDF (S9, 472 responses).}
    \label{fig:s9_robot_types}
\end{figure}

\textbf{Robot types (S9, 472 responses):} 
The types of robots modeled in URDF are shown in \cref{fig:s9_robot_types}.
Of these, single robotic arms were the most commonly modeled (81\%), while delta robots were the least modeled (4\%).
Other types of robots were used by 6\% of the participants, 14\% of whom modeled underwater vehicles.
The low percentage of delta robots modeled with URDF can potentially be related to the limitations of URDF not supporting parallel linkages or closed-chain systems.

Responses to multiple questions provide consistent information.
The fact that URDFs from Universal Robots (54\%) and Kuka (37\%) are most commonly used, and the main types being single robotic arms (81\%) and mobile robots (64\%), implies that these robots are commonly used in the dominant application for URDF which is manufacturing (46\%).

\textbf{URDF and ROS (S10, 472 responses):} 
Looking at the use of URDF and ROS, 66\% stated always using URDF in combination with ROS, 18\% always with ROS but would prefer not to require ROS, and 16\% without ROS.
As described earlier in \cref{sec:methodology}, this question was biased as the users could not choose with and without ROS.
We can only conclude that URDF is commonly used with ROS, by the large number of responses to this (66\%), but also by looking at the high usage of simulation tools compatible with ROS (Gazebo and RViz).
We can also state that a number of the participants prefer URDF's independence from ROS, but we cannot rely on the percentage here, as the question is leading and could affect the responses.


\subsection{Creating URDFs}

\textbf{Tools to create URDF (S11, 441 responses):}
Tools the respondents were aware of are xacro (82\%), SolidWorks URDF exporter (57\%), Fusion2URDF (17\%), OnShape to URDF exporter (12\%), the Blender extension Phobos (12\%), and PTC Creo to URDF exporter (4\%).
Additionally, 90\% have used xacro to generate URDFs (S12, 361 responses).

\textbf{URDF and 3D polygon meshes (S13, 472 responses):}
URDF can be used with basic shapes, such as boxes and cylinders, or with 3D polygon meshes that visualize the robot, see \cref{fig:cylindrical_vs_kuka}.
Most of the participants (93\%) have used URDF with 3D polygon meshes. 

\textbf{Methods to obtain URDF (S14, 439 responses):}
Of the participants that had used URDF with 3D geometrical meshes, 67\% stated they developed the URDF by hand, 61\% obtained the URDF from a CAD tool, 59\% from a ROS package, 47\% from the website of the robot manufacturer, and 20\% from a simulation tool.
It is important to note that this question was asked with the possibility to answer multiple options, and therefore if a participant has developed a URDF by hand, they may have also used a CAD tool for aligning or working with the meshes.
Furthermore, it is possible that some participants have obtained a URDF and subsequently modified it, therefore choosing the option ``developing the URDF by hand''.

Most robots in ROS packages are described using xacro.
The results from the survey showed that 261 participants have used ROS packages, 313 have used xacro, and in total 203 have both obtained URDFs from ROS packages and at some point in their work used xacro.
As most ROS packages contain xacro files that need to be processed with the xacro preprocessor, we would assume that the number of participants that have obtained URDFs from ROS packages would be similar to the number of participants that have used xacro before.
The fact that only 203/261 have obtained URDFs from ROS packages and also used xacro before shows a discrepancy in our expectations.
This may indicate that either many ROS packages contain URDF files that do not need to be preprocessed with xacro, or a misunderstanding by the participants of what a ROS package is.
Another interesting fact about the results is the excessive use of xacro, where 313/361 respondents stated they have used xacro before.
This indicates that it is a useful tool, but similarly to URDF, it has a number of issues~\cite{Albergo&2022}.

\textbf{Difficulty rating of creating/modifying URDFs (S15, 439 responses):}
The participants that have created/modified URDFs rated the difficulty of this, of whom the majority (57\%) rated it as medium, 32\% experienced it as difficult, 7\% experienced it as easy, 1\% answered ``Do not know'', and 3\% stated they have no experience with this.

Furthermore, one of the participants stated \say{\textit{We have a coffee mug in the lab that says, `DO NOT CHANGE THE URDF!'}}, anecdotally illustrating the perceived complexity of modifying URDFs.

\textbf{Challenges with creating/modifying URDFs (S16, 278 responses):}
Of the respondents, that have created/modified URDFs, the most painful parts of the process were setting up the poses and meshes of the robot (21.2\%), the lack of tooling making it difficult to debug the URDF (18.6\%), and adjusting dynamic and kinematic parameters to be accurate (9.0\%).
The full overview of the categories is shown in \cref{tab:painful_creating_modifying_urdf} in the appendix.


\subsection{Challenges, Improvements, and Future Directions}

\begin{figure}
    \centering
    \includegraphics[width=\columnwidth]{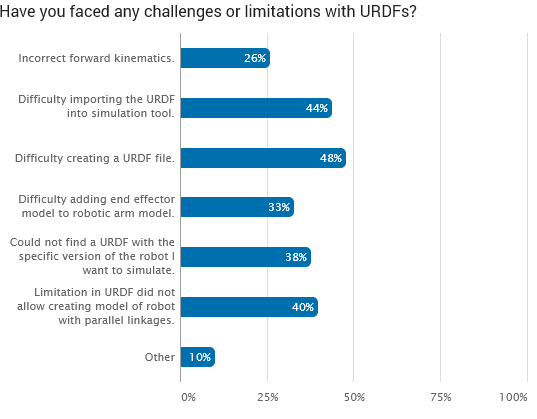}
    \caption{Challenges encountered by the participants (S17, 440 responses).}
    \label{fig:s17_urdf_challenges}
\end{figure}

\textbf{Challenges and limitations (S17, 440 responses):}
The challenges experienced by the participants are shown in \cref{fig:s17_urdf_challenges}.
In addition, 10\% of the participants experienced other challenges presented in \cref{tab:challenges_other} in the appendix.

The most common challenges found in the survey from responses to various questions are summarized in \cref{tab:summarized_challenges_urdf}.

\begin{table}
\caption{Summarized most common challenges of URDF derived from responses to S16, S17, S18, and S19.}
\label{tab:summarized_challenges_urdf}
\begin{tabular}{p{1.35cm}|p{6.5cm}} \hline
\textbf{Challenges} & \textbf{Details} \\ \hline
 \multirow{9}{*}{\textit{Format}}& XML syntax and boilerplate code. \\ \cline{2-2}
    & Multiple xacro files and very long URDF files make it difficult to find components and keep an overview. \\\cline{2-2}
    & Does not support parallel linkages or closed-chain systems. \\\cline{2-2}
    & No real standard and no versioning, making it difficult to know which features of URDF a simulator supports.\\\cline{2-2}
    & Only supports solid bodies, meaning it is not possible to model soft or deformable objects.\\\cline{2-2}
    & There are not enough dynamic parameters, e.g.\ it is not possible to define the elasticity of objects.\\\cline{2-2}
    & Does not support nonlinear mimic joints.\\\cline{2-2}
    & Not possible to define limits to higher-order variables of joints, e.g.\ acceleration, jerk, etc.\\\cline{2-2}
    & Cannot change parameters online in simulation.\\\hline
\multirow{4}{*}{\parbox{1.35cm}{\textit{Creating URDFs}}} & Meshes: difficult to display colors, file paths for meshes, matching visualization and collision meshes, setting the origins.\\ \cline{2-2}
 & Setting up frames and ensuring they are accurate. \\ \cline{2-2}
 & Tedious workflow: debugging is complicated as you need to reload the URDF into a visualization tool every time you make a change, and you need to use multiple tools (CAD, XML, and visualization) to develop one URDF. \\ \cline{2-2}
 & Adjusting parameters to be accurate, e.g.\ the dynamics or kinematic parameters of inertia, joint limits, and positions need to be accurately defined to represent the real world. \\ \hline
\multirow{4}{*}{\textit{Surroundings}} & Lack of documentation makes the learning curve steep. \\ \cline{2-2}
& Lack of tooling: to create robots out-of-the-box, for debugging, for validating URDFs, intellisense or linter, one framework or tool to be able to create/modify/validate URDFs, or a light preview tool for live editing, and tools for converting between URDF and other formats. \\ \cline{2-2}
& More manufacturers should provide accurate kinematic and dynamic parameters. \\ \cline{2-2}
& Difficult to build applications with URDF. \\ \hline
\end{tabular}
\end{table}

\textbf{Improvements (S18, 461 responses):}
The main improvements chosen by the participants are better tools for combining and manipulating URDF files (70\%), easier methods or tools to create URDF files (69\%), better tools for validating URDF files (65\%), better documentation (56\%), a versioned standard (i.e., URDF v1.0, URDF v1.1, etc.) (35\%), and standardized metadata like originator, date, version, etc.
A number of the participants suggested improvements which are categorized in \cref{tab:improvements_other} in the appendix, with the key points being a clear demand for better tooling for creating, combining and manipulating, and validating URDFs.

\textbf{Future directions (S19, 477 responses):}
Of the respondents, 53\% believed that URDF will be more commonly used in the future with the main arguments being that URDF is the de-facto standard in ROS, and as a lot of tools in ROS rely on it, it would be too difficult to replace URDF with a new format.
34\% were not sure of URDF's future use, and 12\% did not believe it would be used in the future.
Their main arguments were that URDF is too limited and instead other formats that tackle these limitations will take over.
More details on these elaborations are presented in \cref{tab:future_yes}, \cref{tab:future_no}, and \cref{tab:future_dont_know} in the appendix.

The main benefits of URDF perceived by the respondents are summarized in \cref{tab:summarized_benefits_urdf}.

\begin{table}
\caption{Summarized benefits and advantages of URDF derived from responses to S19.}
\label{tab:summarized_benefits_urdf}
\begin{tabular}{p{2cm}|p{6cm}} \hline
\textbf{Benefits} & \textbf{Details} \\ \hline
Open-source & Open-source robotics is becoming more common in the industry, and URDF is supported by some of the main ROS tools e.g.\ RViz and Moveit which are used in open-source robotics.\\ \hline
Interoperable & As URDF is not dependent on one specific simulation tool, but is a portable, manufacturer-independent model, it allows users to simulate in any tool that supports URDF and exchange models within the community. \\ \hline
Accessible & It is simple enough (compared to some other formats) that many non-experts can use it and follow the structure of the robot, also considering it is a human-readable format.\\ \hline
Custom models & It allows developers to create custom models of their robots. \\ \hline
\end{tabular}
\end{table}


\subsection{Analysis}

In this section we analyze responses from multiple questions and check if there are correlations in the results.

\begin{figure}
    \centering
    \includegraphics[width=0.9\columnwidth]{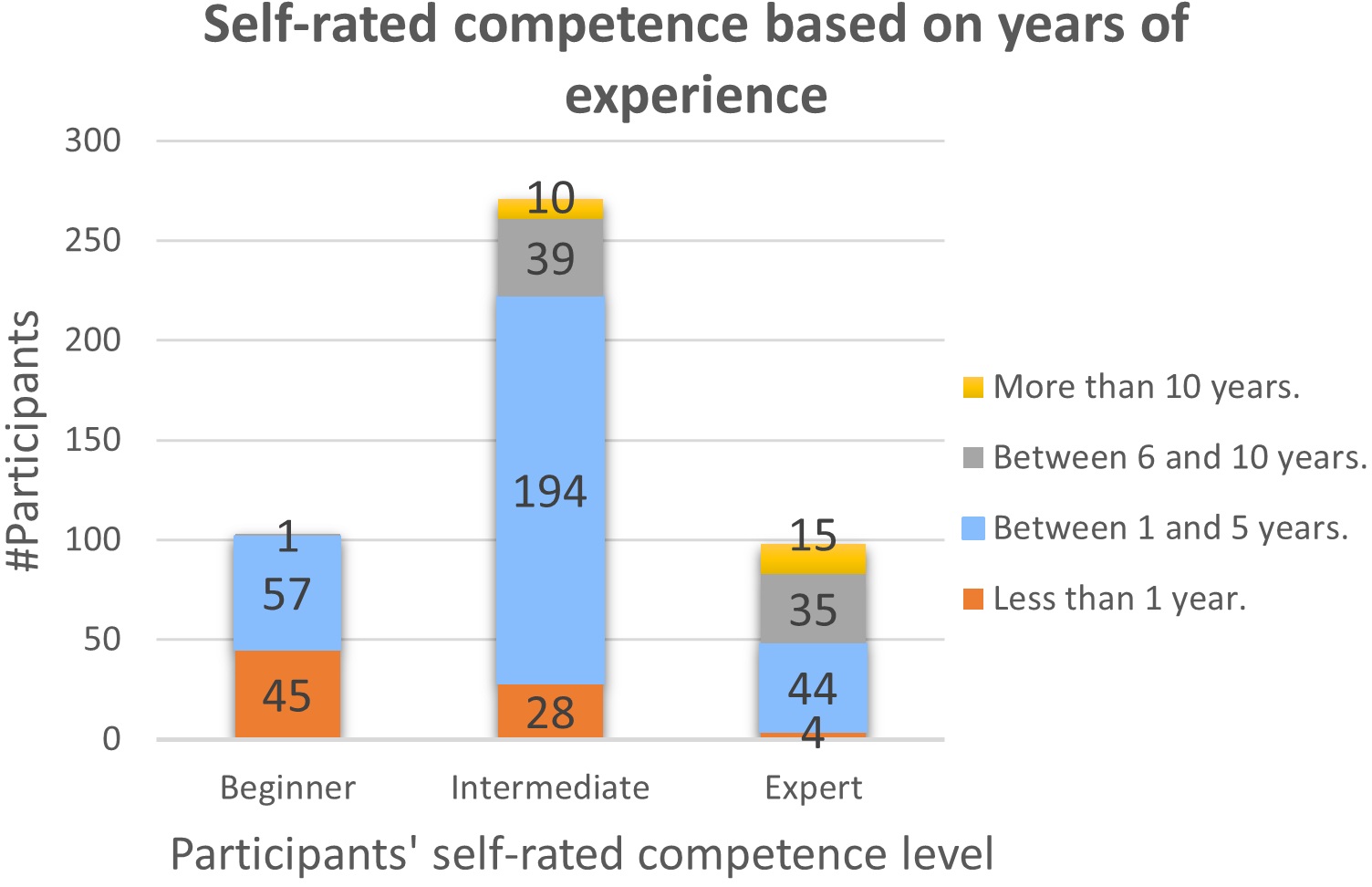}
    \caption{The participants' self-rated URDF competences based on their years of experience with URDF.}
    \label{fig:competence_years_experience}
\end{figure}

To validate the participants' self-rating of their URDF competences, we compare the years of experience with the participants' rating, see \cref{fig:competence_years_experience}.
It is clear that most participants that rate themselves as intermediate have between 1-5 years of experience with URDF, whilst very few with less than a year of experience rate themselves as experts, suggesting the competence self-ratings are reasonable.

\begin{figure}
    \centering
    \includegraphics[width=0.9\columnwidth]{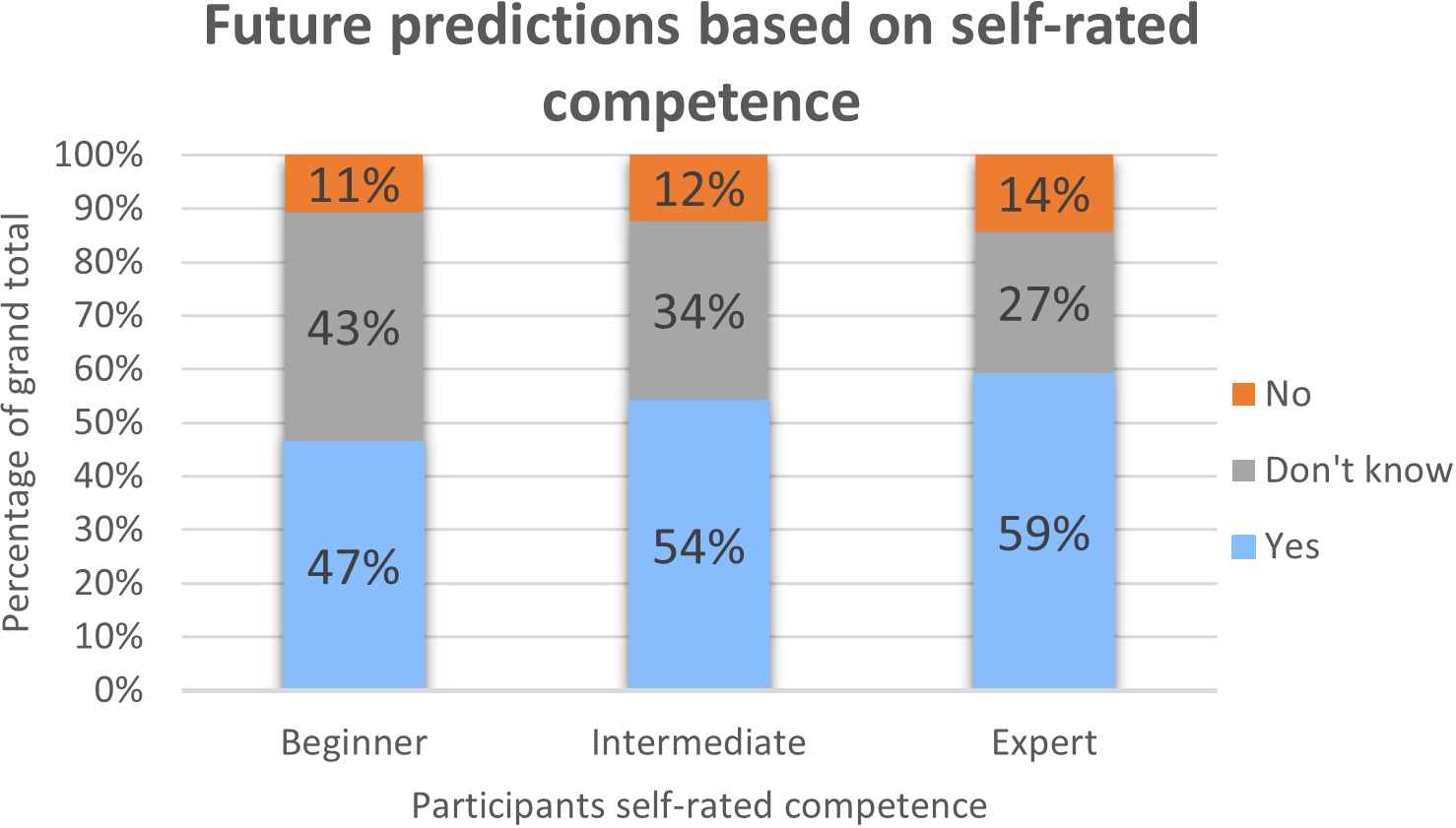}
    \caption{The future predictions of the participants based on their self-rated competences with URDF.} 
    \label{fig:future_predctions_competence}
\end{figure}

\Cref{fig:future_predctions_competence} shows the participants' future predictions based on self-rated competences with URDF.
The results imply there is no significant correlation between the competences of the participants and their future predictions for URDF.

\begin{figure}
    \centering
    \includegraphics[width=0.9\columnwidth]{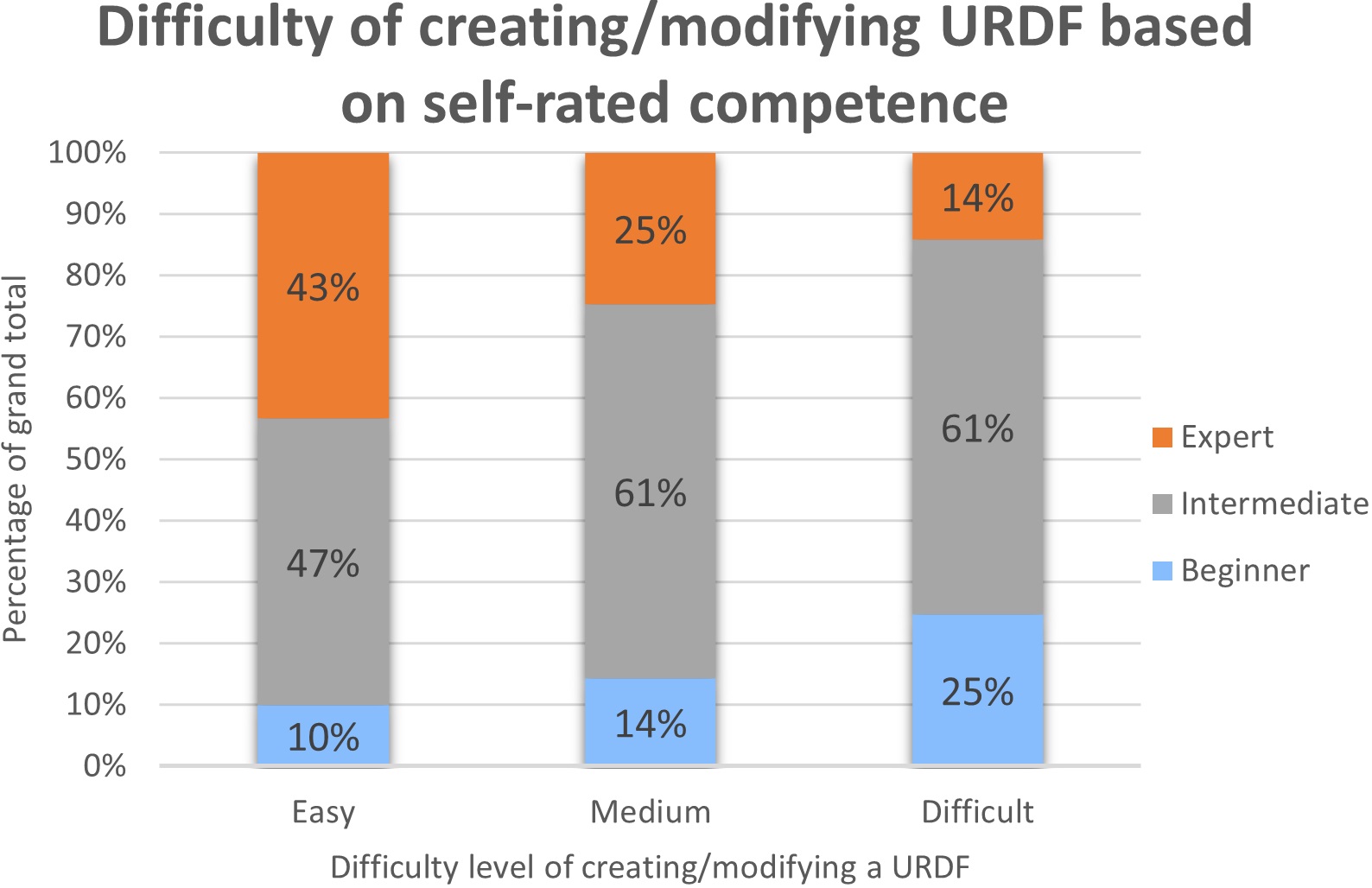}
    \caption{The difficulty of creating/modifying URDFs based on the participants' self-rated competences.}
    \label{fig:difficulty_competence}
\end{figure}

\Cref{fig:difficulty_competence} shows the participants' perceived difficulty of creating/modifying URDFs based on their self-rated competences with URDF.
The results imply that the majority of the participants experienced creating/modifying URDFs as of medium difficulty, regardless of their self-rated competence with URDF.
There is a clear pattern within the beginner and experts groups.
There are fewest beginners in the easy column, with more in medium, and the most in difficult, whilst for the experts the rating is the opposite, i.e., there are fewest experts in the difficult column.
This may indicate, as some of the participants have mentioned themselves, that the learning curve of working with URDF is steep, and with more experience the easier it becomes to work with URDF.

\begin{figure}
    \centering
    \includegraphics[width=\columnwidth]{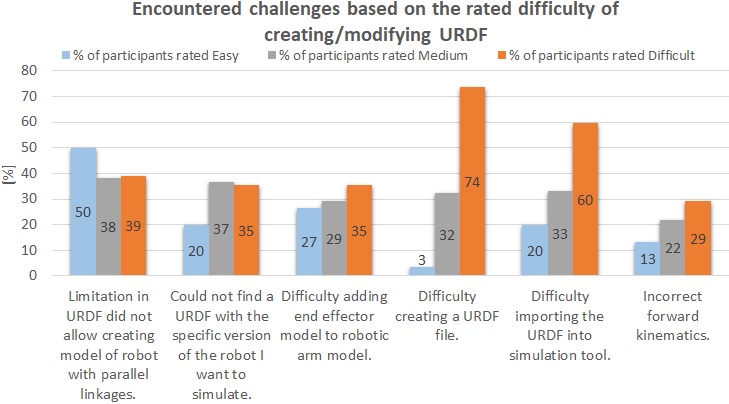}
    \caption{The participants that have chosen a specific difficulty rating for creating/modifying URDFs, and the percentage of the total that have encountered the different challenges.}
    \label{fig:challenges_vs_difficulty}
\end{figure}

Of the participants that found it difficult to create/modify URDFs, 74\% found it difficult to create a URDF (as expected), and 60\% had issues importing URDFs into simulation tools, see \cref{fig:challenges_vs_difficulty}.
Issues with URDF not supporting parallel linkages were experienced by 50\% of the participants that rated the process of creating/modifying URDF as easy, whilst only 3\% of them had issues with creating URDFs.
These results can be associated with the results in \cref{fig:difficulty_competence}, where the majority of the URDF experts rated the process of creating/modifying the file as easy, indicating the learning curve of working with URDF is steep.
This could potentially be a symptom of a lack of documentation.

\begin{figure}
    \centering
    \includegraphics[width=\columnwidth]{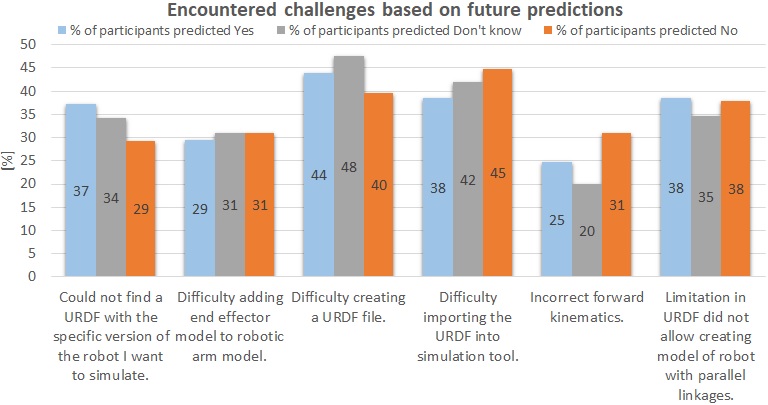}
    \caption{The participants that have provided future predictions of URDF, and the percentage of them that have encountered different challenges.}
    \label{fig:challenges_future_predictions}
\end{figure}

Of the participants that do not believe URDF will be used in the future, 45\% of them have had difficulties importing the URDF into simulation tools, and 40\% have had difficulties creating a URDF, as shown in \cref{fig:challenges_future_predictions}.
Although those numbers are high, there is also a high percentage of participants that believe URDF will be used in the future and have encountered the same challenges, indicating there is no specific challenge that has affected the opinion of the participants.

\section{Summary and Discussion} \label{sec:discussion}

In this section, we answer the questions asked in \cref{sec:methodology} based on the survey results.
Furthermore, discussions on the future focus of URDF are presented.

\subsection{Summary}

\subsubsection{Q1: Who is using URDF?}
    URDF has mainly been used in academic (82\%) and industrial (53\%) organizations. 
    The majority of the users (62\%) have 1-5 years of experience, and have used URDF within the last month (67\%) of taking the survey (conducted December 2022).
    More than 66\% of the participants have used URDF in combination with ROS while others have used it without ROS, showing the format is used both within and outside of the ROS ecosystem.

\subsubsection{Q2: What is URDF being used for?} 
    Manufacturing (46\%) and Transportation (24\%) are the dominating application domains for URDF's use. 
    The most common robot types modeled with URDF are single robotic arms (81\%) and mobile robots (64\%), which are also the main types of robots used in manufacturing.
    Most of the participants use URDF with 3D visualizations (93\%).

\subsubsection{Q3: What are the limitations of URDF?} 
    Most of the respondents had created URDFs by hand, and this is where the main limitations were found.
    When creating a URDF multiple tools are needed, such as CAD software for modifying the meshes, a program to write the XML structure, and additional software to visualize and test the URDF.
    Participants described that the debugging process was tedious; every time a parameter in the URDF file was tweaked, the file was loaded into the visualization software (again) and tested.
    Additionally, difficulties when importing URDFs into simulation tools, and the inability to model parallel linkages or closed-chain systems were reported.

    The main improvements for URDF, desired by the participants, are better tools for creating, manipulating, and validating URDFs.
    Improved documentation to decrease the steep learning curve of URDF was also high on the list of desired improvements for URDF.

\subsubsection{Q4: What is the future of URDF?} 
    The majority (53\%) of the participants believed that URDF will be used in the future.
    Their main argument being that URDF is the de-facto standard in ROS where many tools rely on it to function, making it too difficult to replace URDF with another format.
    The remaining participants were either not sure of or did not believe in URDF's future use.
    Their reasoning was primarily based on the current limitations of URDF. 
    They suspect that another format, which addresses these limitations, will take over in the future.

\subsection{Discussion}

\subsubsection{Other formats}
    Many of the participants mentioned other formats such as SDF~\cite{SDFformat}, USD~\cite{USDformat}, and MJCF~\cite{MJCFormat}.
    A few also mentioned they are working on new formats that they believe will replace URDF.
    With the existing and emerging formats that can be used for modeling robots, it can be overwhelming for newcomers in robot modeling to determine which format to use.
    It may be beneficial to create an overview of the capabilities and limitations of the different formats, to allow users to easily determine which format is most suitable for their needs.

\subsubsection{XML}
    Contrasting opinions were revealed on using XML for describing URDF files.
    Some saw an advantage in its human-readability and simplicity, whilst others believed that XML is outdated and the current structure of the format contains a lot of boiler-plate code.

\subsubsection{Tackling challenges}
    One of the main challenges is having difficulties with importing URDFs into simulation tools.
    Although, we have not analyzed the causes of this, we suggest creating guidelines for tools to support URDF.
    These guidelines would potentially make it easier for simulation tools to implement support for URDF, but also provide a consistent user interface across simulation tools, especially if the same function names are used.
    
    Another interesting challenge is that participants could not find URDFs of the specific versions of the robots they were using.
    This could potentially be tackled by creating a URDF database in which it would be possible for users to share URDFs, rate each other's URDFs, describe issues/limitations, and update their URDF with improved versions.

\section{Conclusion} \label{sec:conclusion}

In this article, we conducted a survey of 510 robotics developers to determine the use of URDF, its current challenges, desired improvements, and future outlooks of URDF by the robotics community.
We found that most of the participants that have simulated a robot before have also used URDF.
Additionally, the majority had used URDF within the last month when taking the survey (conducted December 2022).
The majority of the participants believed that URDF will be more commonly used in the future, especially if it is improved.
We identified challenges and limitations of the format, that the participants would like to be tackled or improved.
Furthermore, we analyzed and discussed results and provided suggestions on how some challenges can be tackled.
The results of this survey can be used as guidelines for future research.

\bibliographystyle{IEEEtran}
\bibliography{IEEEabrv,main}

\newpage
\clearpage
\begin{appendices}

The appendices contain the raw results from the survey in the form of diagrams, and the categorized results from open-ended questions.

\section{Questions} \label{app_questions}

\begin{table}[!h]
\caption{Overview of question IDs, questions, and number of responses in the survey.}
\label{tab:questions_ids_responses}
\begin{tabular}{c|p{5.5cm}|c}\hline
\textbf{ID} & \textbf{Question} & \textbf{\#Responses} \\ \hline
D1 & For how long have you used simulation tool(s) to simulate a robot and/or used URDF files?  & 510  \\ \hline
D2 & In which of these types of environments have you used simulation tool(s) to simulate a robot and/or used URDF files?  & 497 \\ \hline
D3 & If you have used simulation tool(s) to simulate a robot and/or used URDF files at any organization, how large was the most recent organization & 493 \\ \hline
S1 & If you have simulated a robot before, when was the last time you did this?  & 497   \\ \hline
S2 & If you have simulated a robot with 3D visualizations, which tools have you used to do this?  & 494   \\ \hline
S3 & Have you heard of (Unified Robot Description Format) URDF before?  & 497  \\ \hline
S4 & If you have used URDFs before, for how long have you been using them?  & 477  \\ \hline
S5 & When was the last time you used a URDF?  & 472   \\ \hline
S6 & How experienced do you feel you are with URDFs?  & 472 \\ \hline
S7 & In which application domains have you used URDFs? & 472 \\ \hline
S8 & Which manufacturers of robots or end effectors have you used URDFs of?  & 472 \\ \hline
S9 & Which types of robots have you used URDFs of? & 472 \\ \hline
S10 & How have you used URDFs? & 472 \\ \hline
S11 & Are you aware of any of the following or other tools for creating URDFs? & 441 \\ \hline
S12 & Have you ever used the tool ‘xacro’ to generate URDFs?  & 361 \\ \hline
S13 & Have you ever used URDF models (xml file + meshes) with geometrical meshes of the robot (i.e., not just boxes or cylinders, etc.)?   & 472 \\ \hline
S14 & How did you obtain the URDF models? Please mention the specific tools. & 439 \\ \hline
S15 & If you have earlier created/modified a URDF model, what level of difficulty would you rate the process?    & 439 \\ \hline
S16 & What would you say is the most painful part of creating/modifying a URDF model? (Optional)   & 278 \\ \hline
S17 & Have you faced any challenges or limitations with URDFs? & 440 \\ \hline
S18 & Are there any improvements you would like for URDF? & 461 \\ \hline
S19 & Do you think that URDF will be more commonly used in the future? Please justify if possible. & 477 \\ \hline
\end{tabular}
\end{table}

\section{Results} \label{app_results}

Important comments on the survey questions:
\begin{itemize}
    \item ``\textit{The question about if the participant uses URDFs in conjunction with ROS is very limiting and induces a problematic bias.}'' - \textbf{Participant A}
    \item ``\textit{The section about improvements the participant would like to see in URDFs makes it very clear what the researchers would like to see the URDFs, and do not necessarily reflect what the participants want, it reads more like 'which of the things we are already doing should we do first'.}'' - \textbf{Participant A}
    \item ``\textit{Note for Q10: I use URDFs both with and without ROS}'' - \textbf{Participant B}
    \item ``\textit{Note for question 10. I have used URDFs both with and without ROS. No preference on the use - it is situationally dependent.}'' - \textbf{Participant C}
    \item ``\textit{(Side note: 'painful' is a leading word in this question, it makes it clear you are looking for a specific result. 'difficult' or 'challenging' would be more fair)}'' - \textbf{Participant D} (part of answer to question 16)
\end{itemize}

\begin{table}
\caption{The most painful parts of creating/modifying URDFs described by the participants (S16).}
\label{tab:painful_creating_modifying_urdf}
\begin{tabular}{p{2.6cm}|l|p{4.4cm}}
\hline
\textbf{Most painful part of creating/modifying URDF} & \textbf{\%} &\textbf{Description} \\ \hline
Poses and meshes & 21.2 & Would be nice if the manufacturers provided separate meshes for each link and its origin etc., so they easily could be put together in a URDF file. Ensuring correct file paths to meshes can be annoying. Ensuring the axes are accurate. Difficult to get colors to work in the meshes. And also matching the collision and visual geometries. \\ \hline
Lack of tools and difficult to debug & 18.6 & We need intellisense, linters, single tools to develop, debug, and visualize URDFs while in the making with e.g. a preview tool. A tool for validating the URDF or XML file would be beneficial. Would be nice with tools for live editing. \\ \hline
Adjusting the parameters to be correct (e.g. dynamics, physics) & 9.0 & Setting the values of the dynamics, kinematics correctly, e.g. joint limits or position, inertia, etc. \\ \hline
Workflow & 8.8 & The complete workflow of creating a URDF. \\ \hline
Syntax & 8.8 & The syntax is complicated, and XML is a disadvantage. Includes a lot of boilerplate code that is just repeated, e.g. if a box is created for a link, you need to copy it to other links. \\ \hline
Limitations of parallel links and closed chains & 7.1 & The limitation of being able to model robots with parallel links or closed chains. \\ \hline
No real standard & 3.7 & Changing standards and versions without real versioning, parsers have different ways of interpreting frames e.g. Y axis up or Z-axis up. \\ \hline
Lack of documentation & 3.4 & Not enough documentation or not good enough, i.e. not intuitive. \\ \hline
Multiple xacro files or long URDF files & 2.8 & Following the chain of xacro files for more complicated workcells or robots is tedious. Long files make it difficult to keep an overview or find specific components. \\ \hline
Using multiple software tools to generate a URDF is tedious & 1.1 & Using CAD software and other programs is time-consuming and difficult. \\ \hline
\end{tabular}
\end{table}

\begin{table}
\caption{Other challenges described by the participants. The percentage shown in the table is from the 10\% that answered 'Other' to S17 in the survey.}
\label{tab:challenges_other}
\begin{tabular}{p{2.6cm}|l|p{4.3cm}}
\hline
\textbf{Challenges/limitations} & \textbf{\%} & \textbf{Description} \\ \hline
Issues with dynamics & 10.4 & Difficulty creating correct inertial parameters, lack of dynamic parameters. \\ \hline
Complex linkages  & 8.3 & Difficult to model conveyor belts, difficulty modeling more complex linkages even for serial robots. \\ \hline
Not possible to define limits of higher-order variables to joints & 8.3 & Higher-order variable limits cannot be specified such as acceleration and jerk, can also not define coupled joints.  \\ \hline
Lack of tooling & 8.3 & Need tools to create robots out of the box, lack of intellisense, need tools to verify URDF parameters for application. \\ \hline
URDF only supports solid bodies & 6.3 & Lack of support for deformable or elastic links, i.e. limitations modeling non-rigid structures. \\ \hline
Issues with mimic joints & 6.3 & Only linear joint mimic relationships. \\ \hline
No .xsd (standard) & 4.2 & No working/accepted XSD, incomplete format. \\ \hline
Challenges when mixing with SDF  & 4.2 & Turning URDF to SDF is difficult. \\ \hline
Inadequate documentation & 4.2  & Lack of documentation, and need to read through xacro before being able to configure robot. \\ \hline
Issues with origins & 4.2 & Discrepancy between link origin and geometric mesh origin, missing frame support. \\ \hline
Maintainability  & 2.1 & Difficult to maintain. \\ \hline
\end{tabular}
\end{table}

\begin{table}
\caption{Other improvements described by the participants. The percentage shown in the table is from the 11\% that answered 'Other' to S18.}
\label{tab:improvements_other}
\begin{tabular}{p{2.4cm}|l|p{4.5cm}}
\hline
\textbf{Improvements} & \textbf{\%} & \textbf{Description} \\ \hline
Better tools & 19.6 & Syntax checking, linter, online combining and manipulation of URDFs, tools for building and visualizing URDF. \\ \hline
Closed chain support & 14.3 & Support for closed chain robots such as delta robots. \\ \hline
Support for more complex and parallel linkages & 12.5 & Support for complex robots and parallel links. \\ \hline
Improve towards SDF & 8.9 & Better tools to describe scenes, deprecate URDF or change URDF towards SDF. \\ \hline
Standardization  & 5.4 & Standardization, standard method to describe extensions to the format (e.g. new joint types), versioning. \\ \hline
Support for deformable bodies & 3.6 & Deformable links. \\ \hline
Better Gazebo integration & 3.6 & Better integration using Gazebo plugins. \\ \hline
More dynamic parameters & 3.6 & Additional dynamic parameters such as elasticity. \\ \hline
Better conversion support to other formats & 1.8 & Better conversion to other formats such as URDF to USD. \\ \hline
More accurate kinematics/dynamics  & 1.8 & More manufacturers should provide accurate kinematic and dynamic parameters. \\ \hline
\end{tabular}
\end{table}

\begin{table}
\caption{The summarized opinions of the participants that answered YES to URDF being more commonly used in the future (S19).}
\label{tab:future_yes}
\begin{tabular}{p{2.5cm}|l|p{4.4cm}}
\hline
\textbf{Future development thoughts (YES)} &\textbf{\%} &\textbf{Description} \\ \hline
Commonly used/have experience with it already & 19.2 & It is a widely accepted format and many tools are built around it (especially ROS tools). \\ \hline
De-facto ROS standard & 14.7 &  It is the de-facto standard in ROS, and as ROS is a growing community, URDF and its use will grow with it. \\ \hline
Good option &  13.6 &  Good option (best in some opinions) as it is simple to use, enables interoperability, and there seems to be no other alternative. \\ \hline
Good basis, but needs improvement & 10.7 &  It's a great tool but needs usability and workflow improvements and more tools around it. It should also support closed chained and nonlinear mimic joints. \\ \hline
Not too difficult to use/intuitive & 6.8 & It is accessible for non-experts, easy to use, human-readable format, and most developers can use it without being lost in complexity. \\ \hline
Such an interchangeable standard is needed & 6.8 & URDF provides interoperability for robot models, allows people to create models of custom robots, is manufacturer-independent, and there should always be a standard for this. \\ \hline
SOTA robotic simulation tools support URDF & 5.1 & Many software tools and especially robotics software tools are using it. \\ \hline
Increase in use of simulators, open source, and robots overall & 4.0 & More people are working with Open Source Robotics and more companies are valuing simulation before deployment, and with this growth, more people will be using URDF. \\ \hline
Don't know other good formats & 3.4 & It's the best current option for visualization when combined with ROS/moveit/Rviz and there does not seem to be an alternative. \\ \hline
\end{tabular}
\end{table}

\begin{table}
\caption{The summarized opinions of the participants that answered NO to URDF being more commonly used in the future (S19).}
\label{tab:future_no}
\begin{tabular}{p{2.5cm}|l|p{4.5cm}}
\hline
\textbf{Future development thoughts (NO)} & \textbf{\%} &  \textbf{Description} \\ \hline
New standard or other software tools will take over, e.g. SDF, USD, MJCF & 37.1 & A new format or software tool will take over, which will be more robust and easier to use. The format or tool could be SDF, Omniverse, MuJoCo, or USD. \\ \hline
URDF is limited & 12.9 & URDF is limited in terms of support for mechanisms such as closed chain systems, description of actuators and sensors, and has poor abstractions. XML is limiting. \\ \hline
Working with URDF is difficult & 5.7 & It is widely adopted, but people don't seem to like it as it can be formatted in a cleaner way, there is no standardization and few updates. \\ \hline
URDF is not supported by enough software tools & 2.9 & Several simulators do not use URDF. \\ \hline
\end{tabular}
\end{table}

\begin{table}
\caption{The summarized opinions of the participants that answered DON'T KNOW to URDF being more commonly used in the future (S19).}
\label{tab:future_dont_know}
\begin{tabular}{p{2.5cm}|l|p{4.5cm}}
\hline
\textbf{Future development thoughts (DON'T KNOW)} & \textbf{\%} & \textbf{Description} \\ \hline
Other formats may emerge/take over & 24.7 & There are other formats such as SDF that overcome some of the disadvantages of URDF and might be replacing URDF in the future. \\ \hline
URDF is limited & 8.2 & URDF has limitations with regards to modeling systems with more intricate dynamics, not allowing parallel linkages, and has a steep learning curve. \\ \hline
URDF tooling needs to improve & 7.1 & There is a need for tools to address the current limitations and better tools for creating URDFs. \\ \hline
URDF needs improvement to be more commonly used & 7.1 & For URDF to be more commonly used it needs to be improved with a special focus on the creation of URDFs of robots and also building applications using URDFs. \\ \hline
URDF is currently commonly used & 5.9 & URDF is commonly used in the ROS community and is used between different simulation tools. \\ \hline
URDF is difficult to use & 5.9 & URDF has a steep learning curve and limited resources, making it difficult to use and create customized robots. \\ \hline
Such a format is useful & 5.9 & It is important to have a format that is as simple and portable as the URDF. \\ \hline
Depending on the use of ROS & 3.5 & If ROS becomes more widely used then URDF will most likely be as well. \\ \hline
\end{tabular}
\end{table}

\newpage

\begin{figure}
    \centering
    \includegraphics[width=\columnwidth]{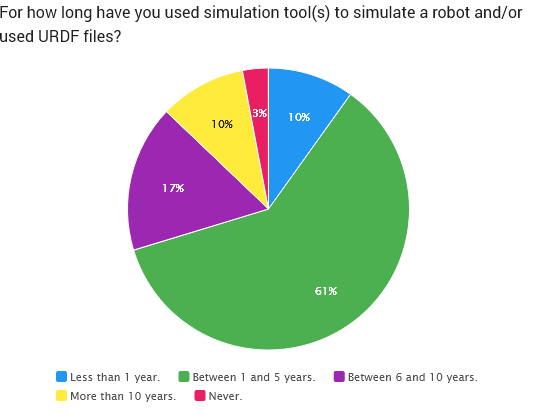}
    \caption{Responses to D1.}
    \label{fig:app_d1_years_simulated_robots}
\end{figure}

\begin{figure}
    \centering
    \includegraphics[width=\columnwidth]{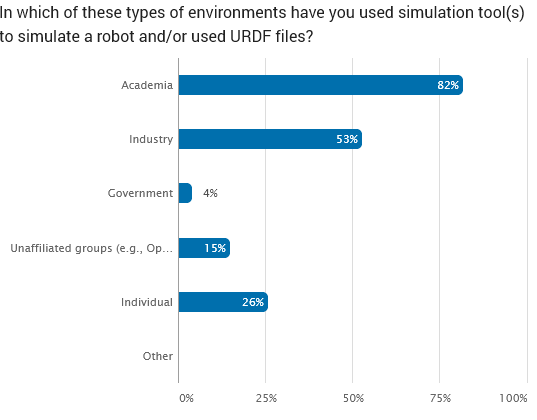}
    \caption{Responses to D2.}
    \label{fig:app_d2_organization_type}
\end{figure}

\begin{figure}
    \centering
    \includegraphics[width=\columnwidth]{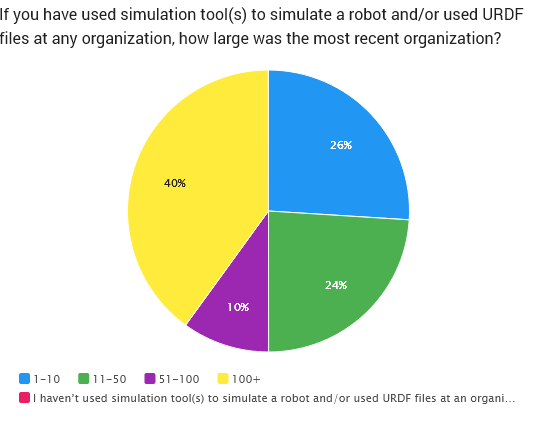}
    \caption{Responses to D3.}
    \label{fig:app_d3_organization_size}
\end{figure}

\begin{figure}
    \centering
    \includegraphics[width=\columnwidth]{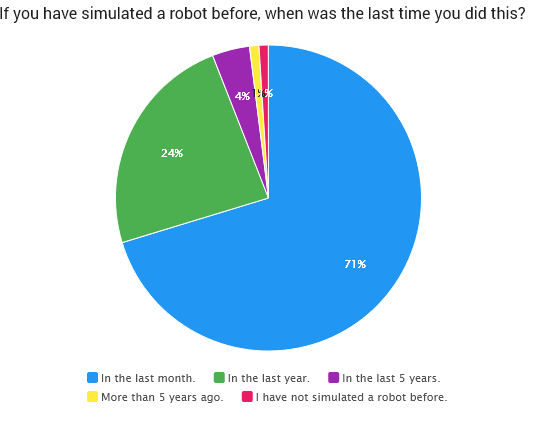}
    \caption{Responses to S1.}
    \label{fig:app_s1_simulate_robot_last_time}
\end{figure}

\begin{figure}
    \centering
    \includegraphics[width=\columnwidth]{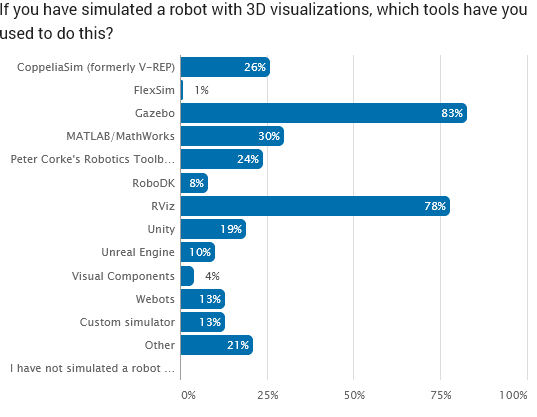}
    \caption{Responses to S2.}
    \label{fig:app_s2_tools_robot_3d_simulation}
\end{figure}

\begin{figure}
    \centering
    \includegraphics[width=\columnwidth]{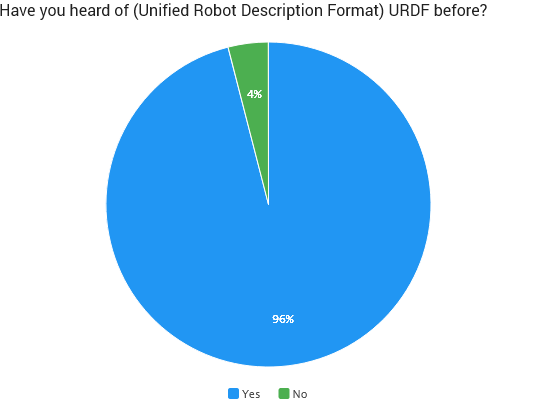}
    \caption{Responses to S3.}
    \label{fig:app_s3_heard_of_urdf}
\end{figure}

\begin{figure}
    \centering
    \includegraphics[width=\columnwidth]{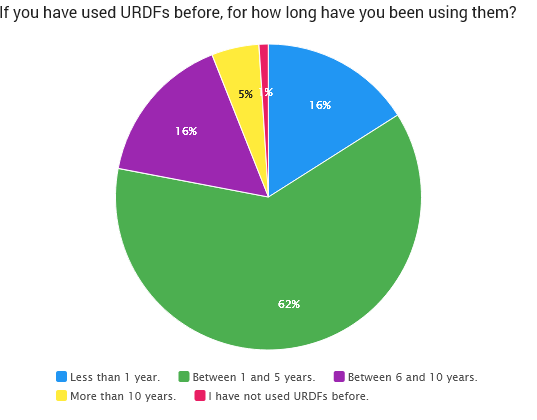}
    \caption{Responses to S4.}
    \label{fig:app_s4_urdf_years_experience}
\end{figure}

\begin{figure}
    \centering
    \includegraphics[width=\columnwidth]{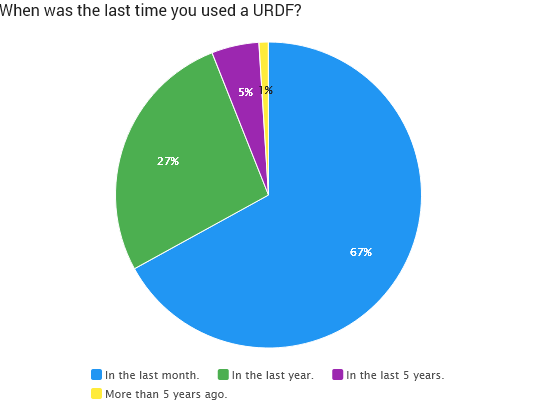}
    \caption{Responses to S5.}
    \label{fig:app_s5_last_urdf_use}
\end{figure}

\begin{figure}
    \centering
    \includegraphics[width=\columnwidth]{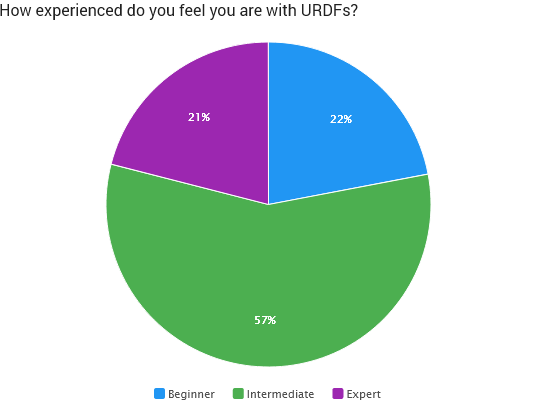}
    \caption{Responses to S6.}
    \label{fig:app_s6_urdf_experience}
\end{figure}

\begin{figure}
    \centering
    \includegraphics[width=\columnwidth]{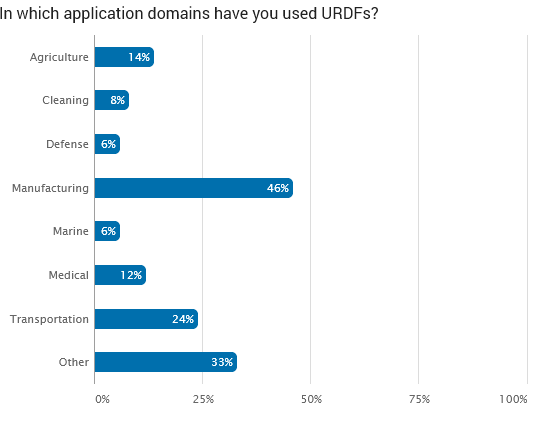}
    \caption{Responses to S7.}
    \label{fig:app_s7_application_domains}
\end{figure}

\begin{figure}
    \centering
    \includegraphics[width=\columnwidth]{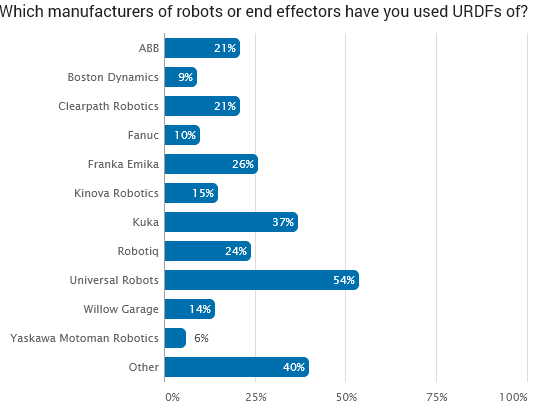}
    \caption{Responses to S8.}
    \label{fig:app_s8_manufacturers}
\end{figure}

\begin{figure}
    \centering
    \includegraphics[width=\columnwidth]{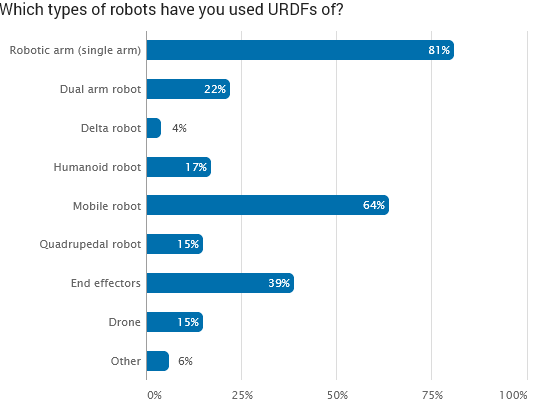}
    \caption{Responses to S9.}
    \label{fig:app_s9_robot_types}
\end{figure}

\begin{figure}[!htbp]
    \centering
    \includegraphics[width=\columnwidth]{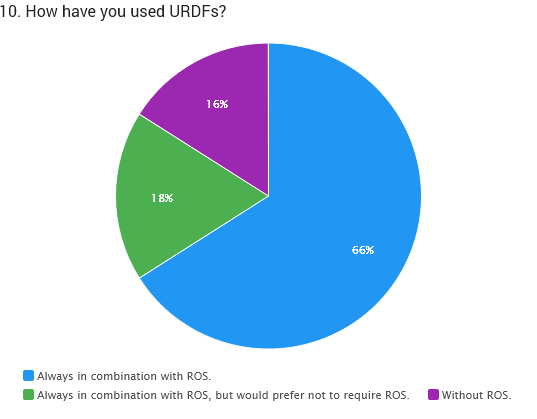}
    \caption{Responses to S10.}
    \label{fig:app_s10_use_urdfs_ros}
\end{figure}

\begin{figure}
    \centering
    \includegraphics[width=\columnwidth]{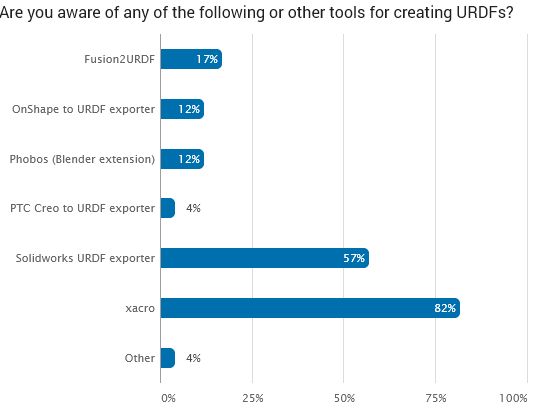}
    \caption{Responses to S11.}
    \label{fig:app_s11_tools_to_create_urdf}
\end{figure}

\begin{figure}
    \centering
    \includegraphics[width=\columnwidth]{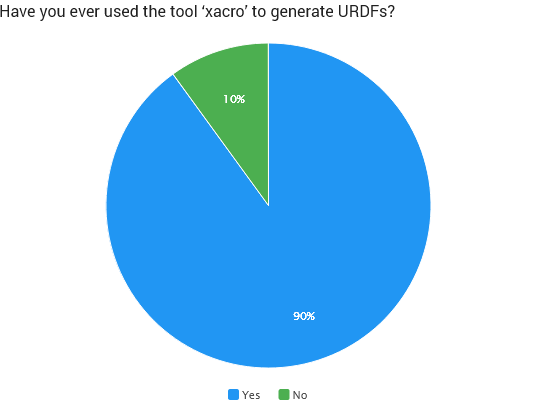}
    \caption{Responses to S12.}
    \label{fig:app_s12_xacro_urdf}
\end{figure}

\begin{figure}[!htbp]
    \centering
    \includegraphics[width=\columnwidth]{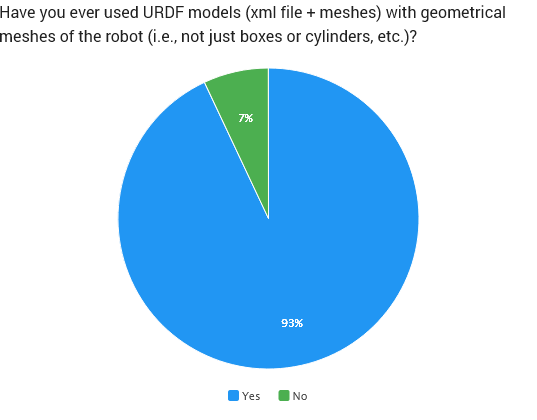}
    \caption{Responses to S13.}
    \label{fig:app_s13_urdf_model_mesh}
\end{figure}

\begin{figure}
    \centering
    \includegraphics[width=\columnwidth]{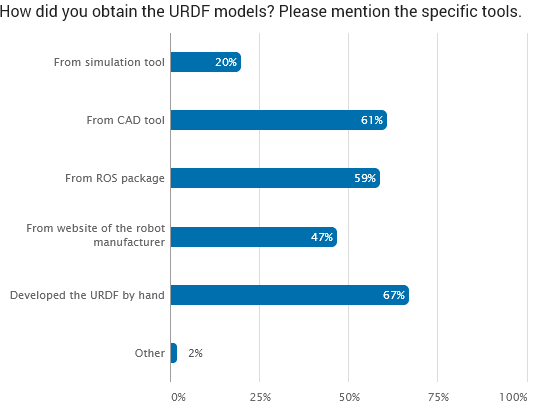}
    \caption{Responses to S14.}
    \label{fig:app_s14_method_obtain_urdf}
\end{figure}

\begin{figure}
    \centering
    \includegraphics[width=\columnwidth]{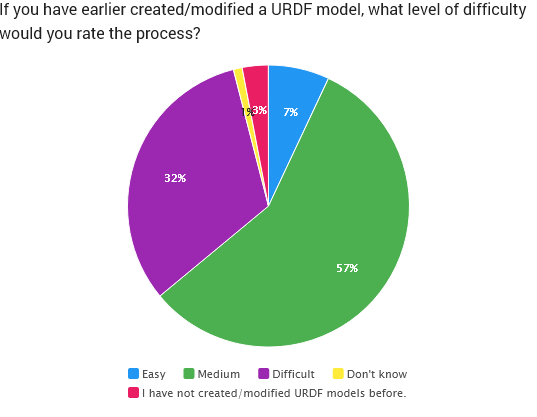}
    \caption{Responses to S15.}
    \label{fig:app_s15_urdf_creation_difficult}
\end{figure}

\begin{figure}
    \centering
    \includegraphics[width=\columnwidth]{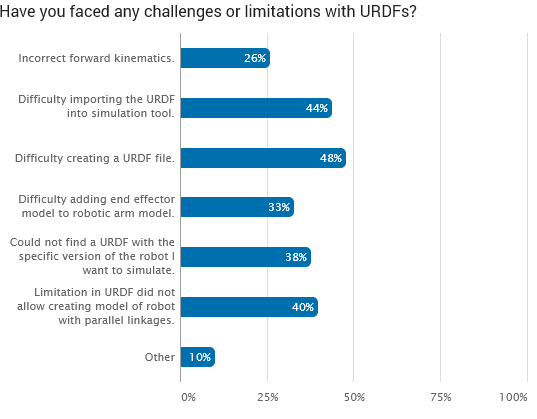}
    \caption{Responses to S17.}
    \label{fig:app_s17_urdf_challenges}
\end{figure}

\begin{figure}
    \centering
    \includegraphics[width=\columnwidth]{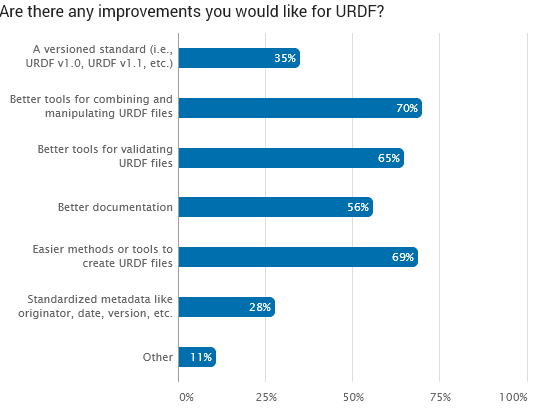}
    \caption{Responses to S18.}
    \label{fig:app_s18_urdf_improvements}
\end{figure}

\begin{figure}
    \centering
    \includegraphics[width=\columnwidth]{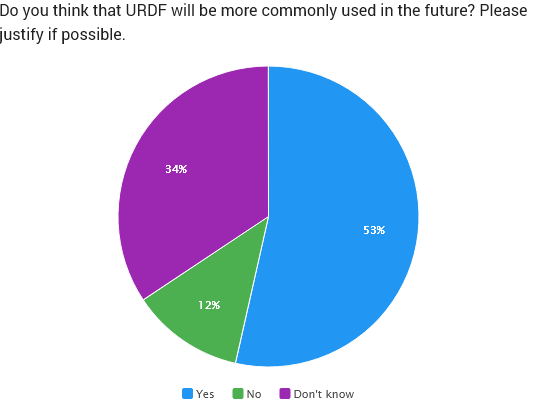}
    \caption{Responses to S19.}
    \label{fig:app_s19_urdf_use_future}
\end{figure}

\end{appendices}

\end{document}